# EndoFSA: Endoscopic Few-Shot Image Generation via Rank-Constrained Parameter Adaptation

Panagiota Gatoula
*Dept. of Computer Science and Biomedical Informatics*
*University of Thessaly*
Lamia, Greece
pgatoula@uth.gr

Grigoris Karypidis
*Dept. of Computer Science and Biomedical Informatics*
*University of Thessaly*
Lamia, Greece
grkarypidis@uth.gr

Dimitris K. Iakovidis
*Dept. of Computer Science and Biomedical Informatics*
*University of Thessaly*
Lamia, Greece
diakovidis@uth.gr

***Abstract*—Wireless Capsule Endoscopy produces large-scale gastrointestinal image data, yet pathological findings remain significantly underrepresented, limiting the generalization performance of deep-learning based abnormality detection systems. Synthetic data generation methods offer a practical solution to mitigate this imbalance. However, their training directly on scarce abnormal samples often results in instability, overfitting and structural distortions. Addressing these challenges requires controlled adaptation mechanisms that preserve anatomical priors, while enabling realistic pathological variation. This paper presents EndoFSA, a GAN-based model for Endoscopic Few-Shot image generation by Adaptation in WCE imaging. EndoFSA leverages a generator pretrained on abundant normal data and adapts it to abnormal domains using limited number of training samples through a rank-constrained parameter adaptation, where only a small number of modulation parameters is updated while the pretrained weights remain frozen. By restricting parameter updates to a low dimensional subspace and incorporating perceptual boundary regularization and cluster-wise diversity control, EndoFSA enables efficient model adaptation under limited data conditions and mitigates mode collapse, while preserving the anatomical priors learned from normal data. Importantly, EndoFSA operates without requiring pixel-level annotations, masks or bounding box supervision. Evaluation on publicly available WCE benchmark datasets spanning various abnormal categories demonstrates that EndoFSA generates abnormal images reproducing real lesions morphology. Moreover, in a downstream classification task, training an image classifier solely on synthetic abnormal images generated by EndoFSA yields performance comparable to that obtained with real images. These findings highlight the effectiveness of few-shot adaptation based on rank-constrained parameter updates for abnormal WCE image synthesis under scarce data conditions.**

## I. INTRODUCTION

Gastrointestinal tract diseases are a leading cause of morbidity and mortality, often requiring early detection of pathological conditions such as inflammation, vascular lesions and polypoid growth for their effective treatment [1]. Wireless Capsule Endoscopy (WCE) [2] provides a non-invasive means of visualizing the entire GI tract, capturing videos comprising thousands of frames per patient. Although this prospect offers comprehensive inspection, manual review of the GI recordings remains a time-consuming process, often requiring more than 60 minutes per examination and is still susceptible to human error even among experienced clinicians.

Deep learning image-based Clinical Decision Support Systems (CDSSs) have shown significant potential in automatic detection of abnormalities in WCE imaging [3], [4]. However, their performance strongly depends on the availability, diversity and representativeness of the training data used for deep learning. Furthermore, for the research community to improve the performance of such CDSSs, it is required that training data be annotated by experts and publicly released for benchmarking. Today such WCE datasets are limited, often imbalanced, vary in size or lack sufficient representation of abnormality categories [5]. This data scarcity limits the generalization of CDSSs, renders abnormality detection a challenging task and constitutes a barrier in the progress in this domain.

Traditional image augmentation techniques, such as rotation, translation or scaling, could be considered as a mitigation measure; however, they usually provide only marginal improvements in the performance of CDSS, as they cannot introduce new pathological patterns [6]. Generative Adversarial Networks (GANs) [7] offer a promising alternative in this aspect by synthesizing realistic abnormal images to enrich the representation of the existing datasets. In the non-medical image domain, GANs have demonstrated remarkable performance for high quality image synthesis, with efficient generation of new unseen data from noise, in a single step [8]. However, in the context of WCE, where abnormal data are scarce compared to the normal ones, training GANs directly on limited abnormal samples is prone to instability, mode collapse and generation of unrealistic artifacts [6]. Most current WCE image synthesis approaches based on GANs either assume sufficient abnormal data availability [9], [10], or focus on specific lesion types [11]. Therefore, few-shot abnormal image generation across diverse categories remains relatively under explored.

To address this limitation, this paper presents Endoscopic Few-Shot image genration by Adaptation (EndoFSA a GAN-based framework for few-shot abnormal image generation . EndoFSA leverages a GAN model [8] pretrained on abundant normal images, and is capable of adapting to the abnormal image domain, using only a limited number of abnormal samples. To this end, EndoFSA generates abnormal images by using rank-constrained parameter adaptation [12], where only a small set of parameters is updated. Unlike the conventional

This work is part of the European project SEARCH, which is supported by the Innovative Health Initiative Joint Undertaking (IHI JU) under grant agreement No. 101172997. The JU receives support from the European Union's Horizon Europe research and innovation programme and COCIR, EFPIA, Europa Bio, MedTech Europe, Vaccines Europe, Medical Values GmbH, Corsano Health BV, Syntheticus AG, Maggioli SpA, Motilent Ltd, Ubitech Ltd, Hemex Benelux, Hellenic Healthcare Group, German Oncology Center, Byte Solutions Unlimited, AdaptIT GmbH. Views and opinions expressed are however those of the author(s) only and do not necessarily reflect those of the aforementioned parties. Neither of the aforementioned parties can be held responsible for them.

fine-tuning strategies, where all the model parameters are optimized, the proposed adaptation strategy constrains parameter updates to a low-dimensional subspace. This approach prevents overfitting or distortion of the previously learned anatomical structures in cases where the available training samples are limited [13], and enables stable adaptation to the abnormal domain, while preserving the structural consistency and anatomical priors learned from normal data. More importantly, the proposed strategy does not require strong supervision, *e.g.*, usage of pixel-level annotations, masks or bounding boxes indicating abnormal regions. Moreover, to further enhance the quality of the synthesis, EndoFSA incorporates perceptual boundary regularization and cluster-wise diversity control [12]. These mechanisms encourage the generation of realistic pathological patterns while preventing mode collapse and maintaining intra-class variation, even under extreme few-shot conditions. In summary, contributions of this study include:

- A GAN-based few-shot model for the generation of abnormal WCE images of various categories, based on the adaptation from normal to abnormal data.
- A rank-constrained model adaptation strategy that enables stable abnormal generation under severe data scarcity, while preserving anatomical structures and without requiring strong supervision.
- A comprehensive evaluation demonstrates the utility of the generated images in downstream classification tasks, alongside quantitative and qualitative assessment of fidelity and diversity.

The rest of this paper is structured in four sections. Section II provides an outline of few-shot image generation strategies with application focus on WCE. Section III details the EndoFSA framework. Section IV describes the experimental setup and reports the results obtained, and the last section summarizes the conclusions derived and draws future research directions.

## II. Related Work

### A. Few-shot image generation

Few-shot image synthesis has been widely studied in various imaging domains, including medical modalities mainly focusing on dermatoscopy images. Regarding non-medical image synthesis tasks, GAN-based models such as Skip Attention GAN (SAGAN) [14] and ProtoGAN [15] have been proposed to preserve structural details under limited training data, using attention mechanisms or prototype-based feature learning strategies, respectively. In the medical imaging domain, dataset limitations render few-shot generation a more challenging task. Thus, only a limited number of studies have been investigated few-shot generation schemes. Recently, MedGAN [16] presented an adaptive altering training strategy based on Wasserstein loss to generate dermatological skin lesions. Nevertheless, the proposed method required a moderate number of data per class.

### B. Endoscopic Image Synthesis

The generation of endoscopic images has proved a challenging task. This is because gastrointestinal tract exhibits variation in textures, illumination, reflections, and appearance of pathological findings. Previous studies [9], [10], [17] applied common GAN architectures like [8] to augment real endoscopic datasets of flexible and WCE imaging modalities, aiming to improve CDSS classification performance. However, these architectures, originally developed for non-medical image synthesis, assume abundant availability of training images. Thereby, their applicability under data constrained settings limits their generation performance. Regarding WCE imaging, the study of [11] presented Sin-GAN-Seg pipeline. An adversarial framework trained with single images that was focusing on polyp lesions. However, the proposed approach relied on per-pixel lesion annotations. In the context of GAN-based endoscopic image generation under limited data, the approach presented in [18] leverages a fine-tuned Image Inpainting GAN to synthesize endoscopic lesions within specified regions of real WCE images. However, the proposed framework relies on predefined lesion locations which limits its ability to replicate detailed and various disease alterations. Although previous studies have investigated WCE image synthesis of abnormal conditions in limited data settings [5], [6], these approaches were focused on alternative generative models such as variational autoencoders, aiming to counterbalance GAN training stability issues that occur with limited training data. Therefore, few-shot image generation of WCE images containing abnormal findings has yet to be thoroughly investigated.

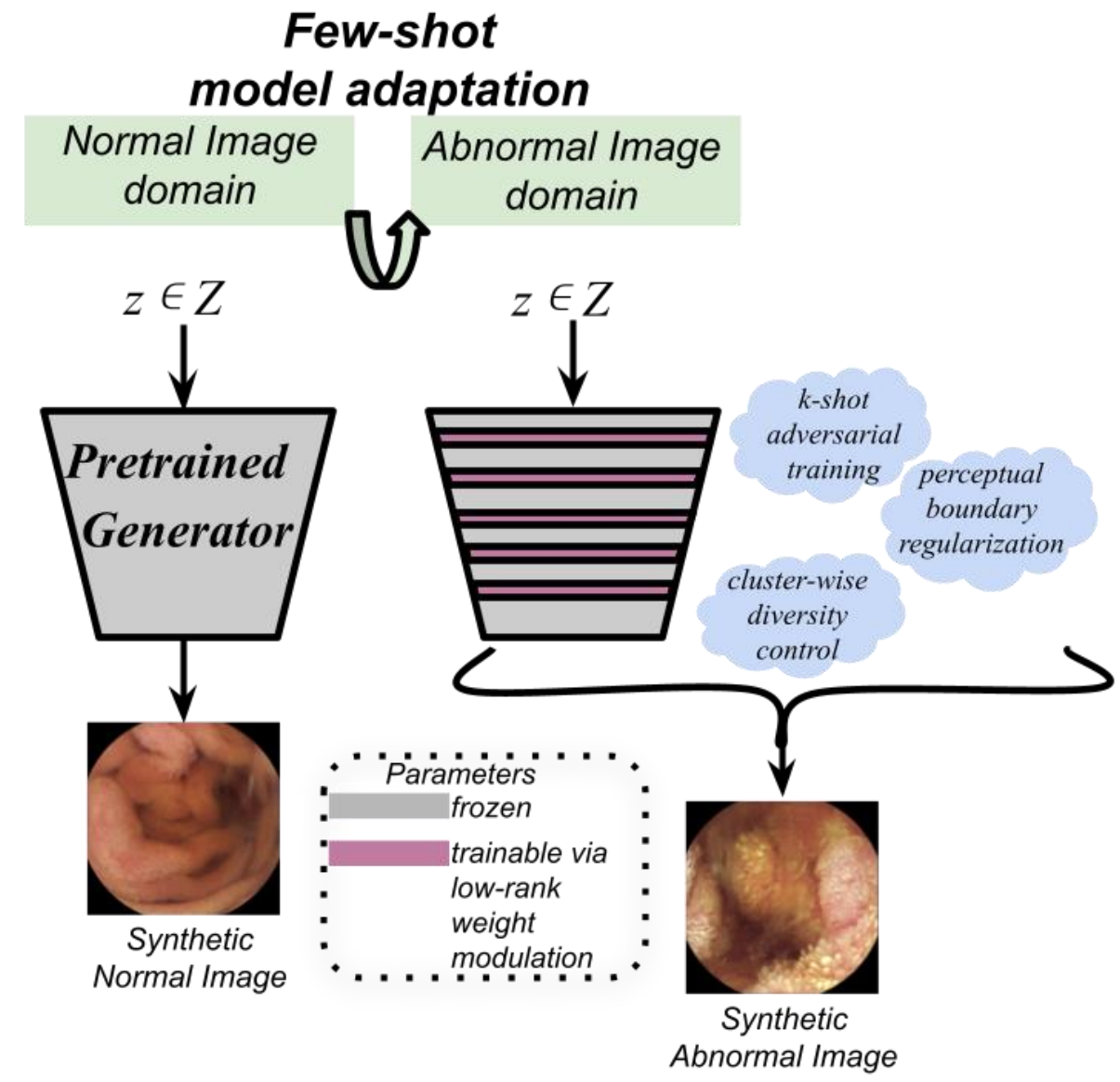

Fig. 1. Overview of the proposed model for few-shot WCE image synthesis of various pathological conditions.

## III. Methodology

This section presents the proposed few-shot image synthesis framework, namely EndoFSA (Endoscopic Few-Shot image genration by Adaptation), for generating WCE images of various pathological cases. An overview of the proposed model is illustrated in Fig. 1. Let $\mathcal{D}_{\text{normal}} = \{x_i^{\text{normal}}\}_{i=1}^{N}$ denote a set of normal images and $\mathcal{D}_{\text{abnormal}} = \{x_j^{\text{abnormal}}\}_{j=1}^{K}$ denote a set of abnormal images, where $K \ll N$. The objective is to synthesize abnormal images that capture clinically relevant pathological characteristics, while preserving the appearance distribution of real abnormal images under data scarcity. The proposed framework leverages a GAN model pretrained on $\mathcal{D}_{\text{normal}}$ and employs constrained few-shot adaptation on $\mathcal{D}_{\text{abnormal}}$ through low-rank weight modulation. Additionally, perceptual boundary regularization and cluster-wise diversity control strategies are employed in the feature space aiming to preserve the anatomical priors learned from $\mathcal{D}_{\text{normal}}$ , while enabling stable adaptation to abnormal data distributions without requiring pixel-level annotations. The methodological components of EndoFSA model are detailed below.

*A. Pretrained synthetic data generation model*

The EndoFSA model adopts StyleGANv2 [8] as a generative backbone. StyleGANv2 is a style-based GAN architecture original proposed for natural image synthesis that has been extensively adopted for medical image generation tasks. Its architecture progressively transforms noise into high-resolution synthetic images, through learned modulation of convolutional features in an intermediate latent space. Let $z \sim \mathcal{N}(0, \boldsymbol{I})$ denote a latent vector sampled from a Gaussian prior distribution. A mapping network $\mathcal{M}: \mathcal{Z} \rightarrow \mathcal{W}$ , composed of fully connected layers, projects $z$ into an intermediate latent representation $w$ defined as:

$$w = \mathcal{M}(z),\ w \in \mathcal{W} \quad (1)$$

The generator network $G_\vartheta$ parametrized by $\vartheta$ , progressively transforms $w$ into an image through a hierarchy of convolutional layers with style-based modulation. For a convolutional layer $\ell$ its weight tensor kernel is defined as:

$$\mathcal{W}^{(\ell)} \in \mathbb{R}^{C_{out} \times C_{in} \times k \times k} \quad (2)$$

where $C_{out}$ and $C_{in}$ represent the number of output and input channels corresponding to layer $\ell$, and $k$ represents the kernel size.

A learned affine transformation $s^{(\ell)} = A^{(\ell)}(\mathcal{W})$ produces channel-wise style coefficients $s^{(\ell)} \in \mathbb{R}^{C_{in}}$. The convolutional weights are modulated by scaling each input channel according to the corresponding style coefficient as follows:

$$\overline{\mathcal{W}}^{(\ell)}{}_{i,j,k} = s_j{}^{(\ell)}\, \mathcal{W}^{(\ell)}{}_{i,j,k} \quad (3)$$

where $i$ and $j$ are the indices of the output and input channels respectively. For preventing signal amplification and ensure variance normalization across channels, weight demodulation is applied. The demodulated weights are computed as follows:

$$\widehat{\mathcal{W}}^{(\ell)}{}_{i,j,k} = \gamma_i{}^{(\ell)}\, \overline{\mathcal{W}}^{(\ell)}{}_{i,j,k} \quad (4)$$

where $\gamma_i{}^{(\ell)}$ represents a demodulation factor for each output channel given by:

$$\gamma_i{}^{(\ell)} = \left(\textstyle\sum_{j,k} (s_j{}^{(\ell)}\, \mathcal{W}^{(\ell)}{}_{i,j,k})^2 + \varepsilon\right)^{-\frac{1}{2}} \quad (5)$$

In (5) constant $\varepsilon > 0$ is used for numerical stability. This modulation-demodulation mechanism decouples style control from feature magnitude, enabling stable training and fine-grained control of image attributes such as textural patterns [8]. In addition, stochastic channel-wise noise injection is applied at each resolution level, to introduce micro-structural variability and enhance synthesis realism. The discriminator network $D_\phi$ , parametrized by $\phi$ , follows a hierarchical convolutional architecture symmetric to the generator, progressively downsampling the input image and producing a scalar realism score. The adversarial objective is formulated as follows:

$$\mathcal{L}_{adv}\left(G_\vartheta, D_\phi\right) = \mathbb{E}_{x \sim p(x)_{real}}\left[\log D_\phi(x)\right] + \mathbb{E}_{z \sim \mathcal{N}(0,\boldsymbol{I})}\left[\log(1 - D_\phi(G_\vartheta(z)))\right] \quad (6)$$

where $\mathbb{E}$ represents the expected value, and $p(x)_{real}$ represents the real data distribution.

The proposed EndoFSA model utilizes a pretrained instance of StyleGANv2 model on $\mathcal{D}_{\text{normal}}$ , thereby encoding typical WCE visual patterns, including anatomical structure, mucosal texture, and illumination statistics. This pretrained prior facilitates stable few-shot adaptation to abnormal cases.

*B. Low-rank weight modulation for few-shot adaptation*

Direct fine tuning of a generator $G_\vartheta$ pretrained on the normal image subset $\mathcal{D}_{\text{normal}}$, using an abnormal image subset $\mathcal{D}_{\text{abnormal}}$, where $K \ll N$, typically leads to rapid overfitting and degradation of previously learned anatomical structures. Aiming to mitigate this issue, in line with [12], the pretrained parameters of $G_\vartheta$ are kept frozen and adaptation is performed through a rank-constrained parameter adaptation mechanism. This significantly reduces the number of trainable parameters while preserving the anatomical knowledge encoded in the pretrained model. The adapted weight tensor $\widetilde{\mathcal{W}}$ (either for a convolutional or a fully connected layer) is defined as:

$$\widetilde{\mathcal{W}} = \mathcal{W}_0 \odot \Delta S + \Delta B \quad (7)$$

where, $\mathcal{W}_0$ denotes the frozen weight parameters (either from a convolutional or a fully connected layer), $\Delta S$ represents scale perturbations, $\Delta B$ represents shift perturbations and $\odot$ denotes Hadamard dot product. The term $\Delta S$ enables the scale modulation controlling the magnitude of the pretrained weights, and the term $\Delta B$ enables the shift modulation allowing adaptation to the abnormal domain. Following [12], both $\Delta S$ and $\Delta B$ are parametrized using rank-one matrices. This formulation allows modulation of the pretrained weights using a minimal number of trainable parameters.

Specifically, for a convolutional layer with weight kernel $\mathcal{W}_0 \in \mathbb{R}^{C_{out} \times C_{in} \times k \times k}$, the pretrained weight tensor $\mathcal{W}_0$ is reshaped into $\mathcal{W}_0 \in \mathbb{R}^{C_{out} \times (C_{in}\, k^2)}$, and the modulation terms $\Delta S$ and $\Delta B$ are parametrized as matrices of rank one as:

$$\Delta S = u_S v_S{}^T \quad (8)$$

$$\Delta B = u_B v_B{}^T \quad (9)$$

where, $u_S, u_B \in \mathbb{R}^{C_{out}}$ and $v_S, v_B \in \mathbb{R}^{C_{in}\, k^2}$. Consequently, the number of trainable parameters is reduced from $C_{out} \times C_{in} \times k^2$ to $C_{out} + C_{in}\, k^2$ for each of the modulation terms. Similarly, for a fully connected layer with a pretrained weight matrix $\mathcal{W}_0{}^{FC} \in \mathbb{R}^{d_{out} \times d_{in}}$, the corresponding $\Delta S$ and $\Delta B$ are defined according to :

$$\Delta S = u_{S_{fc}} v_{S_{fc}}{}^T \quad (10)$$

$$\Delta B = u_{B_{fc}} v_{B_{fc}}{}^T \quad (11)$$

where $u_{S_{fc}}, u_{B_{fc}} \in \mathbb{R}^{d_{out}}$ and $v_{S_{fc}}, v_{B_{fc}} \in \mathbb{R}^{d_{in}}$. In this case, the number of trainable parameters is reduced from $d_{out} \times d_{in}$ to $d_{out} + d_{in}$ per modulation term. The adaptation objective remains adversarial:

$$\mathcal{L}_{adp}\left(\widetilde{G_\vartheta}, D_\psi\right) = \mathcal{L}_{adv}\left(\widetilde{G_\vartheta}, D_\psi\right) \quad (12)$$

where $D_\psi$ represents a discriminator instance trained from scratch, and $\widetilde{G_\vartheta}$ represents the generator with the adapted (modulated) weights.

Let $\vartheta_0$ denote frozen parameters $G_\vartheta$ pretrained on $\mathcal{D}_{\text{normal}}$ subset and let $\varphi$ denote the collection of all rank-one adaptation parameters expressed as:

$$\varphi = \{ u_S, v_S, u_B , v_B, u_{S_{fc}}, v_{S_{fc}}, u_{B_{fc}}, v_{B_{fc}}\} \quad (13)$$

The adapted generator is parametrized as $G_{\vartheta_0,\varphi}$. Few-shot adaptation on $\mathcal{D}_{\text{abnormal}}$ subset is formulated as:

$$\min_{\varphi} \max_{\psi} \mathcal{L}_{adv} \ (G_{\vartheta_0,\varphi}, D_{\psi}) \quad (14)$$

subject to:

$$\nabla_{\vartheta_0} \mathcal{L}_{adv} = 0 \quad (15)$$

Therefore, only the rank-constrained parameters $\varphi$ are updated, while the pretrained model parameters $\vartheta_0$ remain fixed. This ensures preservation of anatomical representations learned from $\mathcal{D}_{\text{normal}}$, while enabling stable adaptation to abnormal cases under limited supervision.

### *C. Perceptual boundary regularization and cluster-wise diversity control*

Few-shot adaptation under extreme sample scarcity (where $K \ll N$), introduces the risks of perceptual drift and memorization of the few abnormal examples by the model. To preserve perceptual consistency, while allowing controlled deviation from the pretrained data manifold of $\mathcal{D}_{\text{normal}}$, a bounded perceptual constraint is imposed in a semantic feature space [12]. This constraint preserves the anatomical structures learned during the generator pretraining on $\mathcal{D}_{\text{normal}}$, while allowing the synthesis of plausible abnormal patterns during adaptation. Let $\Phi(\cdot)$ denote a pretrained perceptual embedding network, such as InceptionV3 [19]. For a synthetic data sample $\tilde{x} = G_{\vartheta_0,\varphi}(z)$ and a real abnormal image $x$ drawn from $\mathcal{D}_{\text{abnormal}}$, the perceptual discrepancy is defined according to:

$$d_{\varphi}(x, \tilde{x}) = \|\Phi(x) - \Phi(\tilde{x})\|_2 \quad (16)$$

where $\|\cdot\|_2$ denotes $L_2$ norm.

Instead of enforcing strict perceptual matching, a boundary-aware constrained is introduced as follows:

$$\mathcal{L}_{per}\left(\widetilde{G_{\vartheta}}, D_{\phi}\right) = \mathbb{E}_{x,z}[\max(0, \ d_{\varphi}(x, \tilde{x}) - \tau)] \quad (17)$$

where $\tau > 0$ defines an admissible perceptual deviation radius. This formulation operates as a regularizing term rather than a reconstruction objective. Only deviations exceeding the radius $\tau$ are penalized, thereby preserving anatomical plausibility, while preventing excessive drift from the prior domain knowledge learned from $\mathcal{D}_{\text{normal}}$, under few-shot conditions [12].

Furthermore, few-shot adversarial adaptation is particularly prone to mode collapse due to the limited $\mathcal{D}_{\text{abnormal}}$ samples. Thus, to encourage diversity while preserving intra-class variability of $\mathcal{D}_{\text{abnormal}}$, a cluster-wise mode-seeking constraint is also considered [12]. This constraint promotes the generation of diverse images for each abnormal pattern, whereas avoiding changes to unrelated anatomical structures. Let the $\mathcal{D}_{\text{abnormal}}$ dataset be partitioned into $C$ clusters, employing *k*-means algorithm in a perceptual feature space defined by a pretrained network $\Phi(\cdot)$, such as a VGG network [20]. Feature vectors $\Phi(x)$ are extracted for each sample $x \sim \mathcal{D}_{\text{abnormal}}$. Then clustering in $C$ groups is performed in the feature space yielding:

$$\mathcal{D}_{abnormal} = \cup_{c=1}^{C} \mathcal{D}_c \quad (18)$$

where each $\mathcal{D}_c$ subset represents a cluster of perceptually similar real abnormal samples. Clustering groups the perceptually similar abnormal samples and enables diversity control within each abnormal pattern, while ensuring sufficient samples per cluster under the few-shot setting [12].

A pair of latent input codes $\left(z_i^{(1)}, z_i^{(2)}\right)$ is sampled from $\mathcal{N}(0, \boldsymbol{I})$ and the images generated from $G_{\vartheta_0,\varphi}$ are assigned to the closest cluster. Within each cluster diversity is enforced by measuring distances of latent-, feature-, and image-space representations with respect to changes between $z_i^{(1)}$ and $z_i^{(2)}$, respectively denoted as $d_{w,i}$, $d_{F,i}$, $d_{\text{I},i}$, and defined in (19)-(21):

$$d_{w,i} = \frac{\left\|w_i^{(1)} - w_i^{(2)}\right\|_1}{\left\|z_i^{(1)} - z_i^{(2)}\right\|_1} \quad (19)$$

where $\|\cdot\|_1$ denotes the $L_1$ norm, and $w_i^{(1)}$ and $w_i^{(2)}$ represent the respective intermediate latent representations defined in (1);

$$d_{F,i} = \frac{1}{L} \sum_{l=1}^{L} \frac{\left\|F_l\,(w_i^{(1)}) - F_l\,(w_i^{(2)})\right\|_1}{\left\|w_i^{(1)} - w_i^{(2)}\right\|_1} \quad (20)$$

where $L$ represents the number of convolution layers considered in $G_{\vartheta_0,\varphi}$, and $F_l\,(w_i^{(1)}), F_l\,(w_i^{(2)})$ their respective feature maps;

$$d_{I,i} = \frac{\left\|G_{\vartheta_0,\varphi}(z_i^{(1)}) - G_{\vartheta_0,\varphi}(z_i^{(2)})\right\|_1}{\left\|w_i^{(1)} - w_i^{(2)}\right\|_1} \quad (21)$$

where $I = G_{\vartheta_0,\varphi}(z)$ represents an image generated by $G_{\vartheta_0,\varphi}$. Thus, the cluster-wise mode seeking regularization term is defined as follows:

$$\mathcal{L}_{cms} = \left(\frac{1}{C} \sum_{i=1}^{C} (\, d_{w,i} + \ d_{F,i} \ + \ d_{I,i})\right)^{-1} \quad (22)$$

Minimization of $\mathcal{L}_{ms}$ encourages the generator to produce noticeable changes in the generated images for different latent codes. This prevents mode collapse while preserving variation within each cluster consisting of perceptually similar abnormal samples. Thus, variability is maintained locally within each abnormal pattern without affecting unrelated anatomical structures. Therefore, the overall adaptation objective is formulated as follows:

$$\min_{\varphi} \max_{\psi} \ (\mathcal{L}_{adv} + \lambda_{per}\mathcal{L}_{per} + \lambda_{cms}\mathcal{L}_{cms}) \quad (23)$$

where coefficients $\lambda_{per}$ and $\lambda_{cms}$ balance perceptual boundary regularization, and diversity preservation, respectively.

## IV. Experiments and Results

### *A. Datasets*

The proposed EndoFSA framework was evaluated on two publicly available WCE datasets, namely KID2 [21] and Kvasir-Capsule [22]. The KID2 video database provides WCE images from the gastrointestinal tract illustrating normal conditions as well as clinically verified abnormalities including inflammatory, vascular, and polypoid findings. The Kvasir-Capsule dataset contains a large collection of annotated WCE frames depicting both normal mucosa and a wide range of pathological findings, including bleeding, ulcers, vascular lesions, and inflammatory lesions. For our experiments, the data from each dataset were divided into two subsets: a subset $\mathcal{D}_{\text{normal}} = \{x_i^{\text{normal}}\}_{i=1}^{N}$, containing exclusively images of normal gastrointestinal mucosa used to pretrain the generator model; and, a subset $\mathcal{D}_{\text{abnormal}} = \{x_j^{\text{abnormal}}\}_{j=1}^{K}$, containing solely images with pathological gastrointestinal findings, used for the few-shot adaptation. In the case of the KID2 dataset, subset $\mathcal{D}_{\text{normal}}$ comprises images from normal small bowel mucosa ($N$=898), and subset $\mathcal{D}_{\text{abnormal}}$ comprises abnormal images ($K$=574) including inflammatory ($n$=227), polypoid ($n$=44) and vascular ($n$=303) lesions. In the case of Kvasir-Capsule dataset, subset $\mathcal{D}_{\text{normal}}$ comprises images from normal small bowel mucosa ($N$=34,338), and subset $\mathcal{D}_{\text{abnormal}}$ comprises abnormal images ($K$=2,873) of angiectasias ($n$=866), erosion ($n$=506), ulcer ($n$=854), lymphangiectasias ($n$=592) and polypoid lesions ($n$=55).

### *B. Experimental Setup*

To simulate realistic clinical scarcity of pathological findings (K << N), few-shot adaptation was performed under three data regimes. For each abnormal class in $\mathcal{D}_{\text{abnormal}}$, $K \in \{10, 50, 100\}$ samples were randomly selected for $K$-shot adaptation of the generator parameters, while the pretrained backbone remained frozen. All images were resized to a fixed spatial resolution of (256×256) pixels and normalized to the range $[-1,1]$.

Initially, the generator network was pretrained on a $\mathcal{D}_{\text{normal}}$ subset, following the experimental settings proposed in [12]. Subsequently, the pretrained instance of generator model was adversarial adapted as described in Section III.B, updating only the rank-constrained parameters, while keeping all pretrained weights fixed. During adaptation process, the discriminator followed the StyleGANv2 [8] hierarchical architecture and it was trained jointly with the generator. For the few-shot adaptation, training settings proposed in [12] were applied. Specifically, the Adam optimizer ($\beta_1 = 0.0, \beta_2 = 0.99$) with a learning rate of $2 \cdot 10^{-3}$ was used. Training was performed for 3,000 iterations with a batch size of 4. Perceptual boundary regularization and cluster-wise mode-seeking were applied with weights $\lambda_{\text{per}} = 0.1$ and $\lambda_{\text{cms}} = 1.0$, respectively. All the experiments were performed on a single NVIDIA GeForce RTX 3090 Ti GPU. The abnormal synthetic images produced were evaluated with respect to their fidelity and utility as described in the following sections.

### *C. Results*

#### *1) Synthetic image utility assessment*

Aiming to assess the utility of synthetic abnormal images generated by the EndoFSA, we evaluated their effectiveness in a downstream classification task. Particularly, a CNN classifier, was trained to distinguish between normal and abnormal WCE images under two conditions: *(i)* using real normal and abnormal images and *(ii)* using real normal images and synthetic abnormal images generated by EndoFSA, which substituted the respective real abnormal images. The performance of trained classifiers was evaluated on held-out test set of real normal and abnormal images, disjoint from those used for the classifier training.

To this end, synthetic abnormal images were generated by EndoFSA model under three few-shot regimes ($K = 10, 50, 100$). Classification experiments were conducted separately for each regime. A widely adopted CNN architecture, InceptionV3 [19] was employed for the classifier training. The CNN was trained using the Adam optimizer with a learning rate of $1 \cdot 10^{-4}$, a batch size of 32 and early stopping to prevent overfitting. The proportions of train, validation and test data splits were 70%, 20% and 10%, respectively. A five-fold cross validation strategy was applied to mitigate potential selection bias. Considering the cluster-wise diversity constraint, a VGG network [20] was used as a feature extractor $\Phi$, and the number of clusters used was $C$=4. The results obtained are summarized in Table I, with bold indicating the best result, whereas underlined text indicating the second-best result. It can be observed that, on the KID dataset, the classifiers trained using synthetic abnormal images generated by EndoFSA achieved a performance that is comparable to that of the classifiers trained with real images, across all regimes. Notably, the synthetic abnormal images generated by EndoFSA achieved the highest AUC, which is 93.94±0.74% for *K=10*, closely approaching classification performance on real images, which yielded an AUC of 95.09±0.98%. It is worth noting that 70% of the generated abnormal images by EndoFSA had a very high similarity (>80%), in terms of utility, with the respective real abnormal images, as assessed using the recently proposed Interpretable Utility Similarity (IUS) measure [23]. In the case of the Kvasir-Capsule dataset, the highest AUC was achieved for *K=100*, resulting in an AUC of 86.62±1.47%. This AUC was lower than that achieved using the respective real abnormal images, which is 97.17±0.37%. This lower performance can be attributed in part to the presence of subtle abnormalities *e.g.*, angiectasia and erosions included in that dataset, which usually tend to be less prominent.

TABLE I. DOWNSTREAM CLASSIFICATION PERFORMANCE ON REAL DATA USING INCEPTIONV3.

| | ***KID dataset*** | | ***Kvasir-Capsule dataset*** | |
|---|---|---|---|---|
| | ***Accuracy*** | ***AUC*** | ***Accuracy*** | ***AUC*** |
| *Real Dataset* | **90.22±0.76** | **95.09±0.98** | **91.29±0.54** | **97.17±0.37** |
| *EndoFSA (K=10)* | <u>87.95±0.75</u> | <u>93.94±0.74</u> | 70.59±1.96 | 77.95±1.11 |
| *EndoFSA (K=50)* | 85.90±1.92 | 91.96±1.38 | 75.93±1.53 | 84.23±1.65 |
| *EndoFSA (K=100)* | 86.75±1.21 | 92.56±0.86 | <u>77.85±2.71</u> | <u>86.62±1.47</u> |

For reference, classification results obtained using synthetic abnormal images generated by EndoFSA trained without few-shot constraints (*i.e.,* using the entire $\mathcal{D}_{\text{abnormal}}$ subsets). These are denoted as EndoFSA (*full*) in Table II. For comparison, the same Table includes results obtained using synthetic abnormal images generated from a conventional StyleGANv2 model [8], trained using the same $\mathcal{D}_{\text{abnormal}}$ subsets. On KID dataset, EndoFSA (*full*) generates abnormal images exhibiting classification performance comparable to StyleGANv2 images, with a marginal difference in terms of AUC. On Kvasir-Capsule dataset, synthetic abnormal images from StyleGANv2 yielded higher AUC compared to EndoFSA (*full*). However, it should be highlighted that StyleGANv2 was trained following conventional adversarial training, updating all model weights, whereas EndoFSA follows a model adaptation training approach, where only a small subset of parameters is trained. Overall, these results demonstrate that EndoFSA generates images with practical utility similar to that of real images in WCE classification tasks, particularly under limited data conditions *i.e.,* using 100 or fewer samples.

#### *2) Quantitative assessment of synthetic image quality*

To further evaluate synthetic abnormal images generated by EndoFSA, two widely adopted synthetic image assessment measures, namely Fréchet inception distance (FID) [24] and Learned Perceptual Image Patch Similarity (LPIPS) [25] were calculated. FID quantifies how realistic synthetic abnormal images appear with respect to real abnormal images. Lower FID scores indicate that the synthetic abnormal images generated by EndoFSA, better represent real image characteristics. LPIPS evaluates the diversity between randomly selected synthetic image pairs, capturing differences in texture and fine-grained structures. Higher LPIPS values indicate more varied synthetic images, which is important under few-shot settings. Table III summarizes FID and LPIPS scores obtained for synthetic EndoFSA images produced by each few-shot regime alongside scores obtained for synthetic images generated by EndoFSA model trained without few-shot constraints (i.e., trained on the whole $\mathcal{D}_{\text{abnormal}}$ subsets) which is denoted as EndoFSA (*full*). Bold indicates the best scores obtained. On KID dataset, the lowest (best) FID score of 75.60 was achieved for *K=50*, whereas on Kvasir-Capsule the lowest FID score of 74.03 was achieved for *K=100*, indicating that for the respective training regimes the abnormal WCE images generated by EndoFSA more closely resemble the real abnormal. It can be noticed that training EndoFSA using the entire $\mathcal{D}_{\text{abnormal}}$ data distribution, resulted in only a marginal FID score improvement of 0.53% for the KID dataset, whereas there was no improvement for Kvasir-Capsule. LPIPS scores remained relatively consistent across all models with only marginal differences of 1.00%, suggesting similar perceptual diversity across different training strategies. For KID dataset the highest LPIPS score of 0.42 was observed for *K=10* and *K=50* regime, whereas for Kvasir-Capsule the highest LPIPS score of 0.53 was observed for *K=10* regime. Overall, these results indicate that EndoFSA exhibits almost consistent performance across various data regimes.

TABLE II. COMPARISON OF DOWNSTREAM CLASSIFICATION PERFORMANCE ON REAL DATA USING INCEPTIONV3 FOR DIFFERENT ADVERSARIAL TRAINING APPROACHES.

| | ***KID dataset*** | | ***Kvasir-Capsule dataset*** | |
|---|---|---|---|---|
| | ***Accuracy*** | ***AUC*** | ***Accuracy*** | ***AUC*** |
| *EndoFSA (full)* | **88.66±2.03** | 94.54±1.01 | 76.22±1.84 | 85.44±2.12 |
| *StyleGANv2* | 87.17±2.68 | **95.11±0.63** | **87.14±0.57** | **94.57±0.55** |

TABLE III. QUANTITATIVE ASSESSMENT OF SYNTHETIC IMAGES

| | ***KID*** | | ***Kvasir-Capsule*** | |
|---|---|---|---|---|
| | ***FID↓*** | ***LPIPS↑*** | ***FID↓*** | ***LPIPS↑*** |
| *EndoFSA (k=10)* | 81.94 | **0.42** | 80.38 | **0.53** |
| *EndoFSA (k=50)* | **75.60** | **0.42** | 75.45 | 0.52 |
| *EndoFSA (k=100)* | 77.39 | 0.41 | **74.03** | 0.52 |
| *EndoFSA (full)* | 76.86 | 0.41 | 76.39 | 0.52 |

### 3) *Qualitative evaluation of synthetic images*

Figure 2 presents a visual comparison between the abnormal images generated by the proposed EndoFSA model and real abnormal images provided by WCE datasets. It can be observed that the synthetic images produced by EndoFSA,

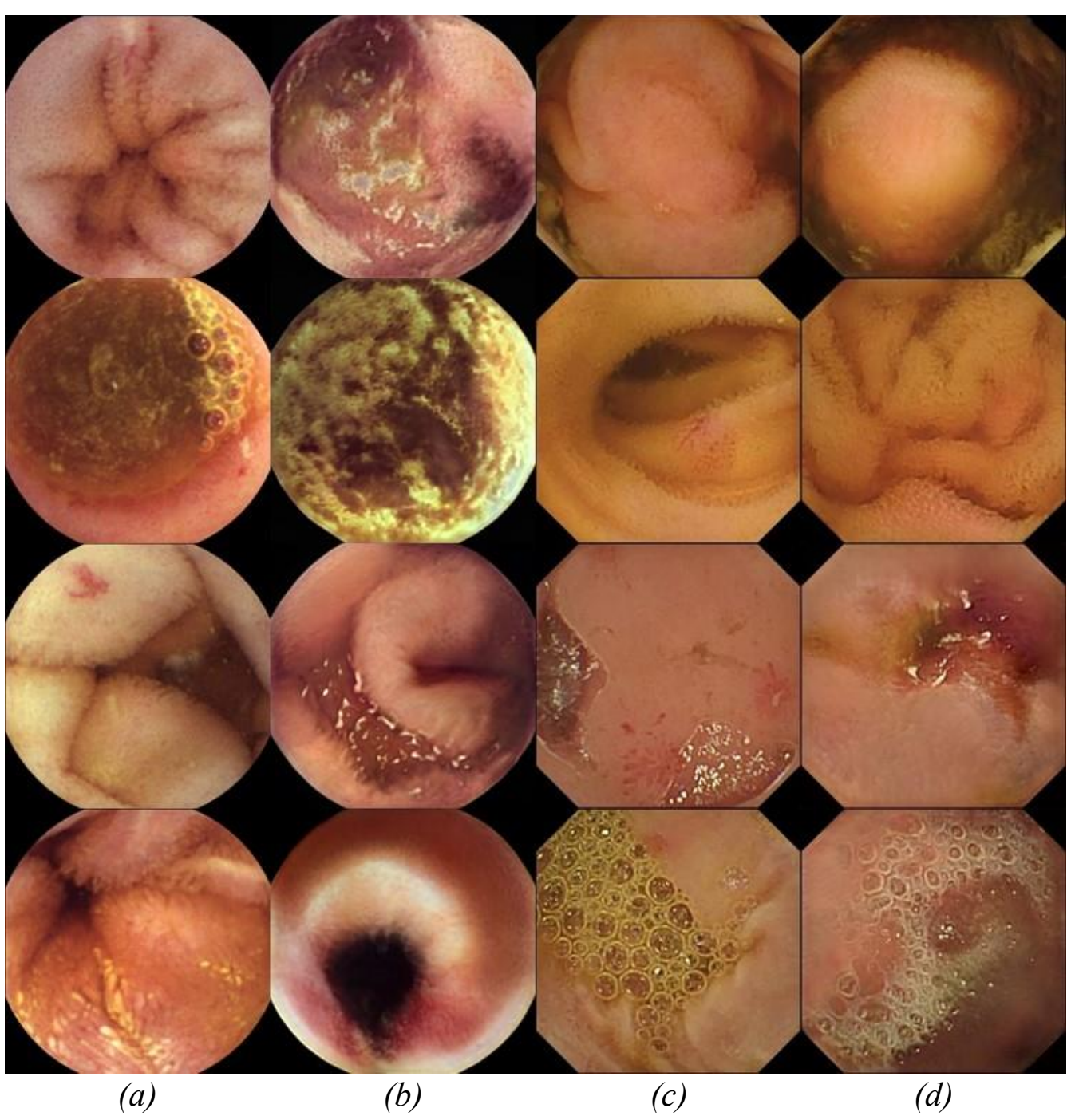

*(a)* *(b)* *(c)* *(d)*

Fig. 2. Real and synthetic WCE images generated by EndoFSA illustrating various abnormal conditions. (a) Real KID images. (b) Synthetic KID images (c) Real Kvasir-Capsule images. (d) Synthetic Kvasir-Capsule images. The synthetic images were generated using *K=10* (first row), *K=50* (second row) and *K=100* (third row) samples from $\mathcal{D}_{\text{abnormal}}$ subset and using the entire (*full*) $\mathcal{D}_{\text{abnormal}}$ subset (fourth row).

closely reproduce the appearance of endoscopic abnormal tissue. In particular, the synthetic images generated by EndoFSA, preserve the visual characteristics of endoscopic lesions demonstrating anticipated color and texture variations as well as lesion morphology. Synthetic abnormal images exhibit lesion appearance resembling that of the real ones, indicating that the cluster-wise diversity control effectively mitigates mode collapse when adversarial training is performed under limited data settings *i.e.,* using 100 or less of training samples. Moreover, it can be noticed that abnormal conditions are integrated in the endoscopic tissue without disturbing the surrounding normal tissue and relevant anatomical structures, demonstrating that the low-rank constrained adaptation training preserves priors learned from normal WCE images. Additionally, the lesion findings appear naturally blended with the normal endoscopic background, while maintaining mucosal characteristics commonly captured during WCE examinations like bubbles, and illumination reflections.

## V. CONCLUSIONS

This paper presented EndoFSA, a GAN model for few-shot WCE abnormal image generation. By leveraging a generator pretrained on abundant normal data and adapting it through rank-constrained parameter updates, the proposed framework enables plausible image generation of abnormal images, using only a limited number of abnormal training samples *i.e.*, 10, 50, 100. By constraining adaptation to low-dimensional subspace, EndoFSA preserves the anatomical priors learned from normal gastrointestinal images, while allowing controlled synthesis of abnormalities without requiring pixel-level annotations or additional supervision. Experimental evaluation across WCE benchmark datasets containing diverse abnormal findings showed that the EndoFSA produces realistic and diverse synthetic samples, as reflected by quantitative metrics and qualitative assessment. Furthermore, training a CNN classifier using exclusively synthetic abnormal images produced by EndoFSA resulted in performance comparable to that achieved with real images. Overall, the results obtained indicate that low-rank constrained adaptation of pretrained GANs provides a practical optimization strategy in the context WCE abnormal image generation under low data regimes. As future research, we aim to extend EndoFSA to simultaneously generate corresponding abnormality masks alongside synthetic images, further improving its applicability for detection and segmentation tasks.